\documentclass[conference]{IEEEtran}
\usepackage{times}

\usepackage[numbers]{natbib}
\usepackage{multicol}
\usepackage[bookmarks=true]{hyperref}

\usepackage{amssymb}
\usepackage{mathtools}
\usepackage{graphicx}

\usepackage{subcaption}
\usepackage{url}
\usepackage{tikz}
\usetikzlibrary{calc}
\usetikzlibrary{positioning}

\usepackage{adjustbox}
\usepackage{array}
\usepackage{makecell}
\usepackage{booktabs}
\usepackage{tabularx}
\usepackage{algorithm}
\usepackage{algpseudocode}
\usepackage[capitalize]{cleveref} 

\begin{document}

\title{Counterfactual Reasoning and Environment Design for Active Preference Learning}

\author{Yi-Shiuan Tung, Bradley Hayes, Alessandro Roncone}

\author{\authorblockN{Yi-Shiuan Tung, Bradley Hayes, and Alessandro Roncone}
\authorblockA{Department of Computer Science, University of Colorado Boulder\\
\{yi-shiuan.tung, bradley.hayes, alessandro.roncone\}@colorado.edu}}

\maketitle

\begin{abstract}

For effective real-world deployment, robots should adapt to human preferences, such as balancing distance, time, and safety in delivery routing. Active preference learning (APL) learns human reward functions by presenting trajectories for ranking. However, existing methods often struggle to explore the full trajectory space and fail to identify informative queries, particularly in long-horizon tasks. We propose CRED, a trajectory generation method for APL that improves reward estimation by jointly optimizing environment design and trajectory selection. CRED ``imagines'' new scenarios through environment design and uses counterfactual reasoning--by sampling rewards from its current belief and asking ``What if this reward were the true preference?''--to generate a diverse and informative set of trajectories for ranking. Experiments in GridWorld and real-world navigation using OpenStreetMap data show that CRED improves reward learning and generalizes effectively across different environments.
\end{abstract}

\IEEEpeerreviewmaketitle

\section{Introduction}
\label{sec:introduction}

Planning under user preferences often involves trade-offs among competing objectives, such as time, cost, and risk. However, users’ preferences over these trade-offs are rarely known ahead of time and can vary across individuals and contexts, making it difficult for autonomous systems to make decisions that align with user expectations. One illustrative domain is autonomous delivery, where robots must select routes that balance factors like travel time, energy consumption, tolls, and surface conditions (e.g., paved vs. unpaved) \citep{10218729, barnesmassively}. For example, a user might accept higher energy costs for substantial time savings (e.g., traversing grass) but not for marginal gains. These preferences are hard to hand-specify and vary between different businesses and individuals \citep{yang2015toward}.

Preference learning (PL) aims to learn a reward function based on human preferences between pairs of trajectories, eliminating the need to manually define rewards and allowing for personalization. Unlike inverse reinforcement learning which requires demonstrations, PL allows non-expert users to provide input in complex domains like Atari games, robot locomotion, and routing \citep{christiano2017deep, knox2011augmenting, wilde2020apl}. However, PL can be sample inefficient, often requiring numerous human preferences for accurate reward learning, which limits its applicability in high-dimensional problems. To address this, Active Preference Learning (APL) finds preference queries—sets of robot trajectories presented to a human—that maximize information gain \citep{Sadigh2017ActivePL, biyik2018batch}. 
Previous work optimizes over a pre-generated set of trajectories \cite{erdem2020asking} or the replay buffer \cite{lee2021pebble} to make optimization tractable, but this method may not sufficiently explore the full feature space, resulting in poor generalization to novel environments, especially for long-horizon problems such as robot routing.

\begin{figure}[t]
    \centering
    \begin{subfigure}{0.49\columnwidth}
        \includegraphics[width=\linewidth]{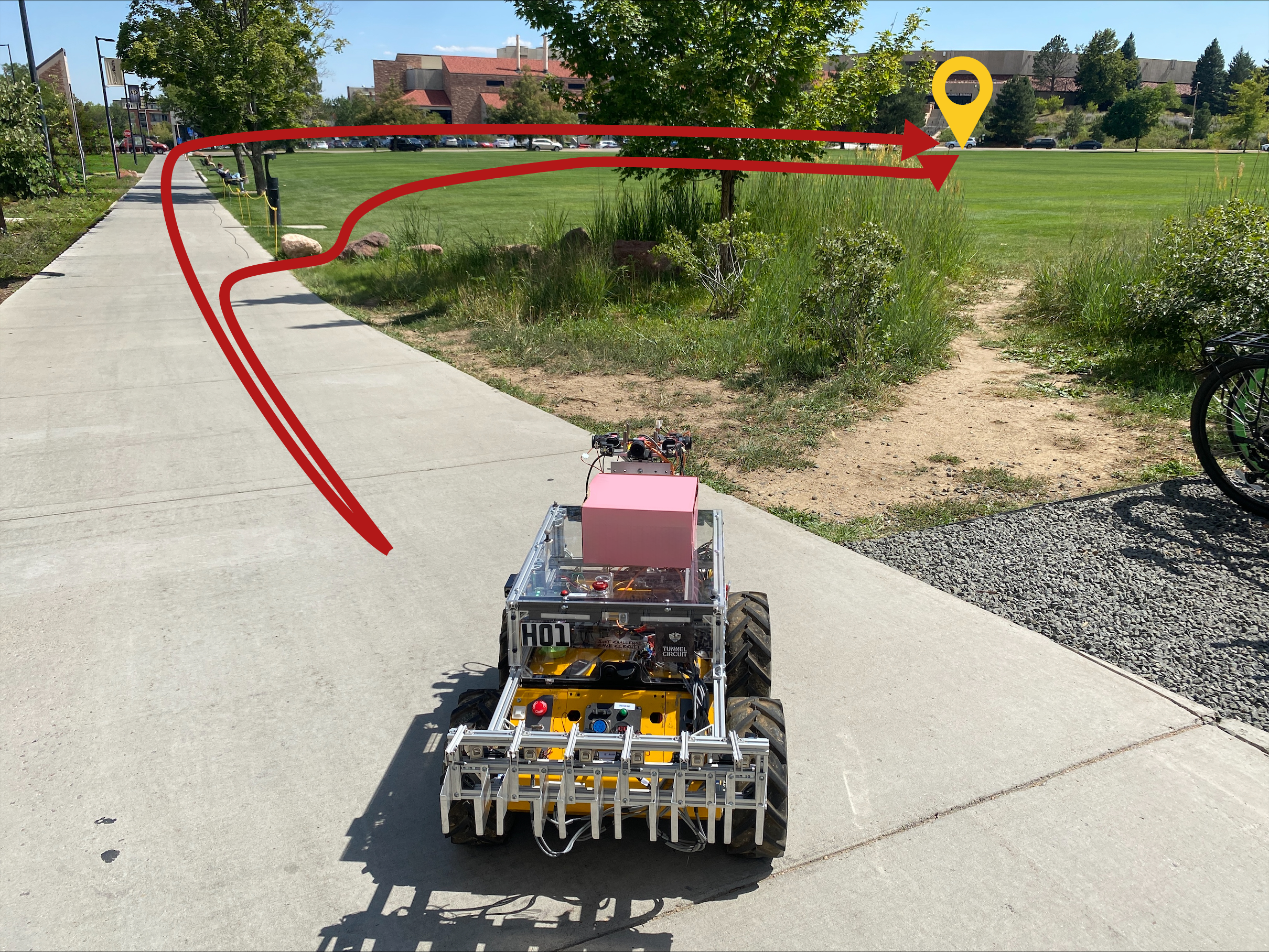} 
        \caption{Suboptimal queries}
    \end{subfigure}\hfill
    \begin{subfigure}{0.49\columnwidth}
        \includegraphics[width=\linewidth]{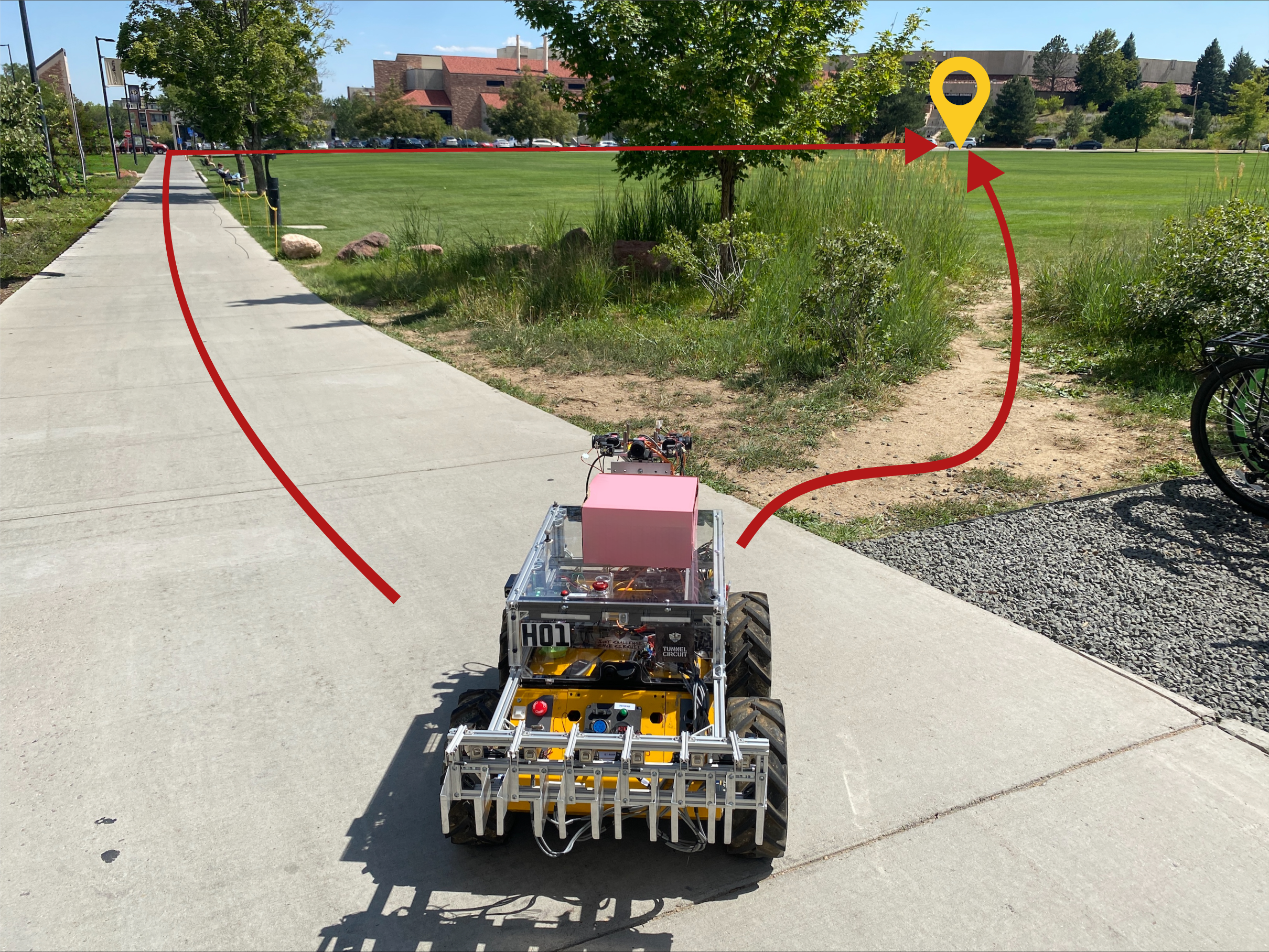} 
        \caption{Counterfactuals}
    \end{subfigure}\hfill
    \begin{subfigure}{0.49\columnwidth}
        \includegraphics[width=\linewidth]{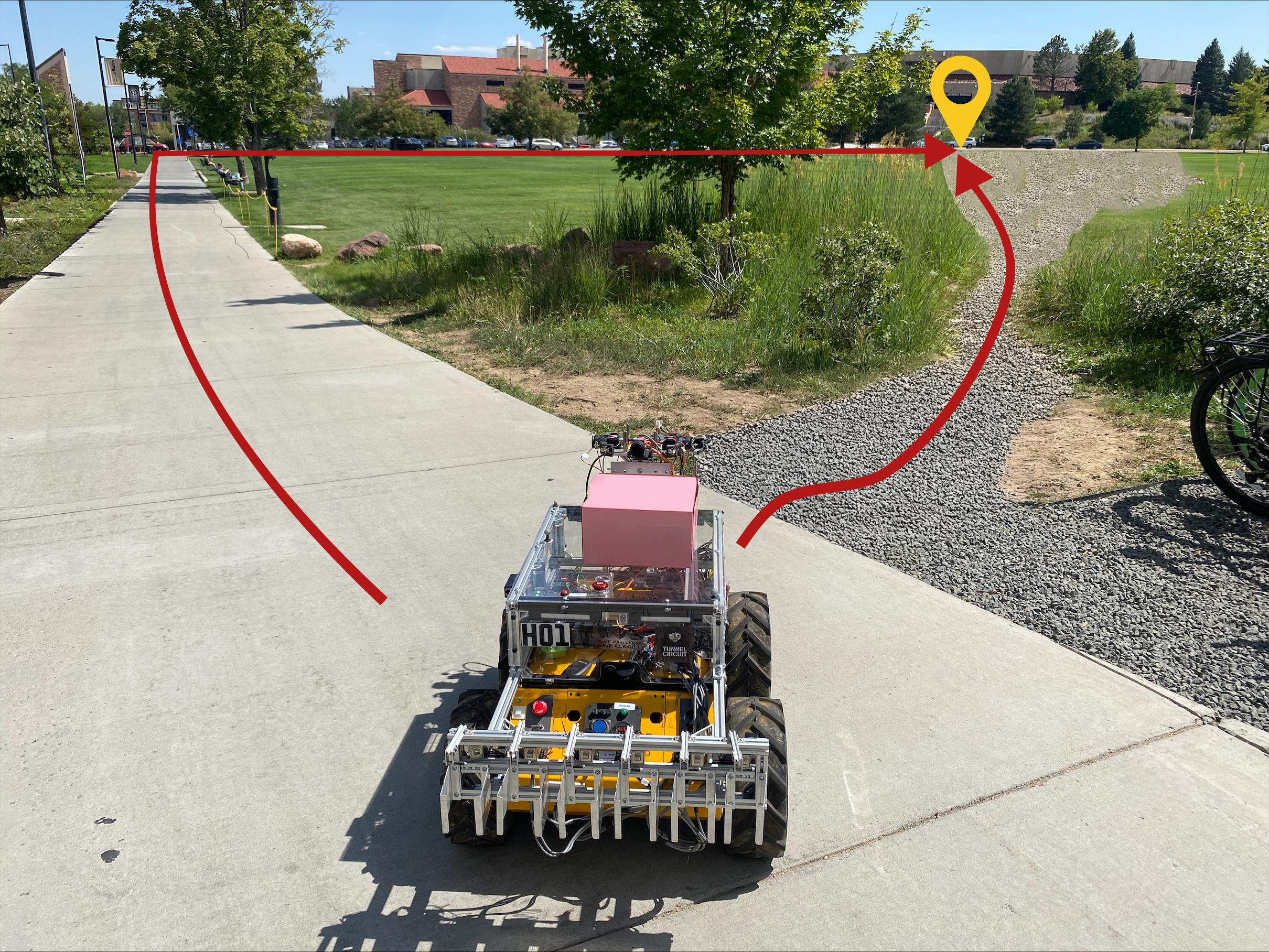} 
        \caption{Environment design}
    \end{subfigure}\hfill
    \begin{subfigure}{0.49\columnwidth}
        \includegraphics[width=\linewidth]{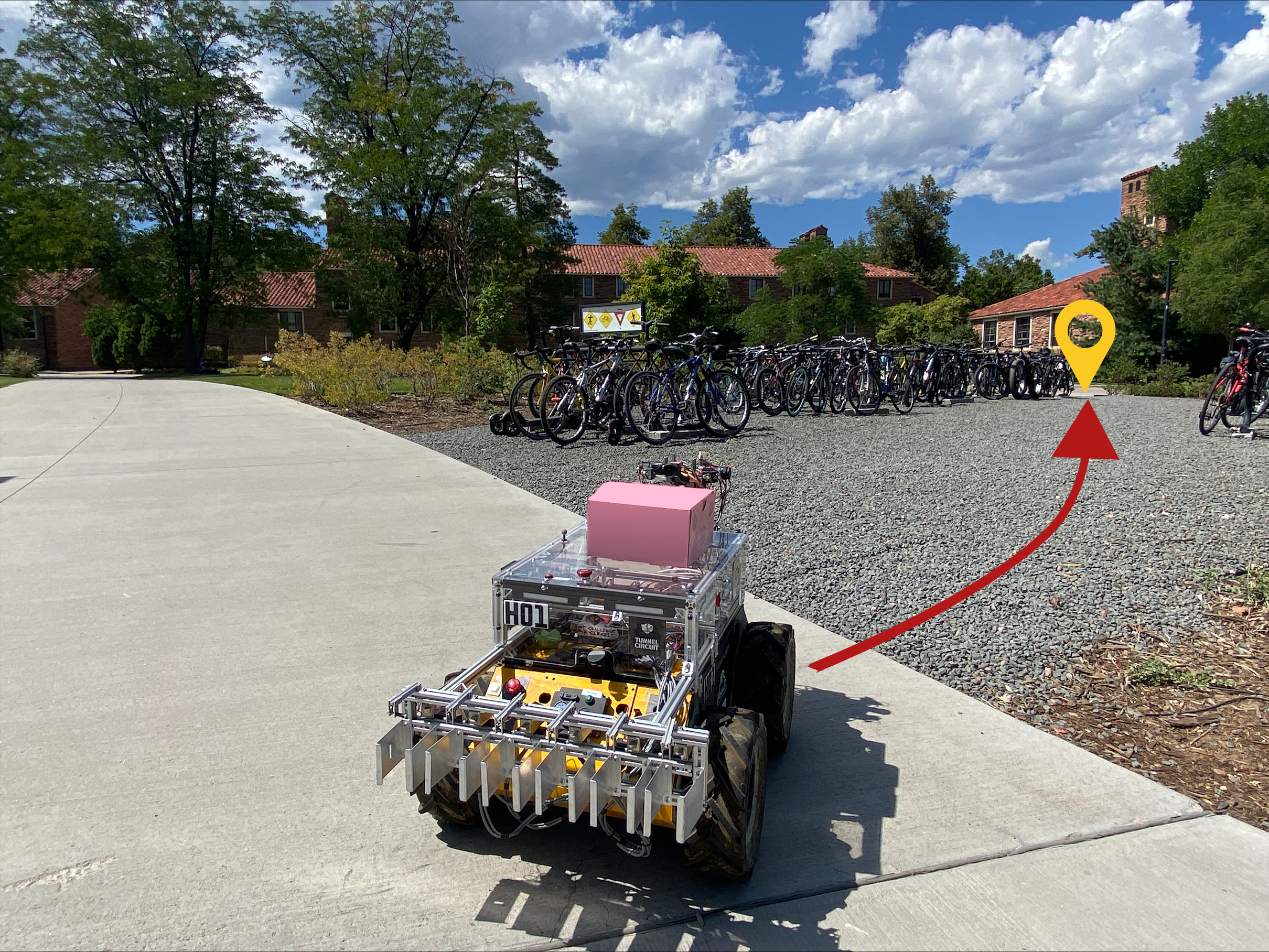} 
        \caption{Generalization}
    \end{subfigure}
    \caption{The delivery robot above (goal: yellow pin) optimizes its path by considering factors like travel time, energy, safety, and surface conditions. Through active preference learning, it infers human rewards from trajectory rankings. (a) However, current state-of-the-art methods often struggle to efficiently generate informative trajectory pairs for these queries, leading to suboptimal results. To overcome this, our approach incorporates two key contributions: (b) counterfactual reasoning, which explores varied hypothetical preferences to produce more diverse trajectories, and (c) environment design, which "imagines" different scenarios—such as altering terrain from grass to gravel—to enhance the system's generalization capabilities. (d) As a result, the robot can more effectively adhere to human preferences, even in novel environments.}
    \label{fig:intro_figure_condensed}
\end{figure}

We introduce CRED, a novel and efficient query generation method for APL that learns reward functions which generalize to different scenarios by using 1) \textbf{C}ounterfactual \textbf{R}easoning and 2) \textbf{E}nvironment \textbf{D}esign. Instead of generating queries through random rollouts \cite{erdem2020asking, lee2021pebble}, CRED's counterfactual reasoning generates queries that reflect different hypothesized human preferences. Assuming a linear human reward model where $R_H(\xi) = w^T\phi(\xi)$ (the dot product of reward weights $w$ and trajectory features $\xi(\phi)$), learning the reward function simplifies to learning $w$. We use Bayesian inference \citep{ramachandran2007bayesian} to maintain a belief over $w$, updated with each human query. By sampling diverse $w$ values (based on cosine similarity) from the current belief, CRED directly evaluates different human preferences and performs counterfactual reasoning, effectively asking "what if $w_i$ or $w_j$ were the true reward weight?" Our second key insight is that the environment influences the generated trajectories and thus the information value of queries. CRED employs Bayesian Optimization \citep{bayesianopt2014} to efficiently find environment parameters that yield the most informative preference queries.

Experiments in a GridWorld domain and real-world navigation using OpenStreetMap data \citep{OpenStreetMap} demonstrate that CRED generates more informative preference queries and converges faster to the true reward function compared to prior methods. Furthermore, CRED learns rewards that generalize to new environments by querying human preferences in "imagined" scenarios. We assume the robot has knowledge of the features and that ground truth rewards generalize; our goal is to learn the reward weights such that they do generalize. In summary, our contributions are: 1) an environment design approach that enables querying human preferences across diverse contexts, 2) a query generation method that uses counterfactual reasoning which generates trajectories that better reflect different human preferences, and 3) a demonstration of our method's ability to generalize to novel environments through experiments in two domains. Our approach enables robots to learn human preferences more effectively and with fewer iterations.

\section{Related Works}
\label{sec:related_works}

\textbf{Robot Routing.} Prior work in autonomous robot routing often employs optimization methods to minimize travel time and costs \citep{ostermeier2023multi}. However, these methods have difficulty adapting to dynamic environments in real-time due to the high computational cost of the optimization problems, prompting the use of reinforcement learning (RL). \citet{bozanta2022courier} and \citet{chen2024optimization} apply RL for route planning but focus on rewarding successful deliveries without considering trade-offs between different factors. In contrast, this paper models the reward as a function of several features such as travel time, distance, terrain types etc. \citet{barnesmassively} incorporates road properties such as distance, surface condition, and road type, and learns a reward function from a large Google Maps dataset. Our approach, however, focuses on personalization, enabling rapid adaptation of learned rewards by iteratively querying the human for preferences and eliminating the need for large demonstration datasets.

\textbf{Preference Learning.}
Preference learning learns the human's reward function by presenting the human with robot trajectories and then asking the human to pick the best one \citep{erdem2020asking}. In active preference learning (APL), the objective is to find the most informative preference query (i.e. a pair of trajectories to query the user) by finding the query that maximizes the expected difference between the prior and the posterior belief distributions over the rewards \citep{Sadigh2017ActivePL}. Subsequent work improves the objective to maximize the mutual information of the query and the estimated weights which generates queries that are easier for humans to answer \citep{erdem2020asking}. To improve the time efficiency of APL, \citet{biyik2018batch} proposes batched queries so that queries can be answered in parallel.

A major challenge of APL is generating trajectories that maximize the mutual information objective. \citet{erdem2020asking} and \citet{lee2021pebble} address this by optimizing within a fixed set of trajectories, but this approach does not scale for long-horizon problems. In addition, the preference queries are generated within the context of the current environment, and previous research has not utilized environment design to improve query quality. Our work uses counterfactual reasoning to generate trajectories based on a belief distribution of reward weights while also optimizing environment parameters to maximize mutual information, thereby improving the information gain of the resulting preference queries.


\textbf{Environment Design in RL and Robotics.}
Environment design treats environment parameters as optimizable variables. In RL, it has been leveraged for curriculum learning to improve generalization and convergence, for instance, through co-evolution of agents and environment difficulty \citep{wang2019paired} or by modifying parameters to maximize an agent's learning potential \citep{dennis2020emergent, azad2023clutr}. In human-robot interaction, environment modification has been used to generate interpretable robot behaviors \citep{kulkarni2020designing}, legible human motion \citep{tung2024workspace, Tung2023HRI}, and to support collaborative teaming in settings like warehouse design \citep{ijcai2023p611} or tabletop reorganization \citep{bansal2020supportive}. Our work applies environment design within APL to find informative preference queries that can effectively reduce the posterior entropy (uncertainty) of the learned reward function.

\section{Preliminaries}
\textbf{Model.} We consider a fully observable environment modeled as a Markov decision process (MDP) consisting of $\{\mathcal{S}, \mathcal{A}, \mathcal{T}, \mathcal{R}, \gamma, \mathcal{S}_0\}$, where $\mathcal{S}$ is the set of states, $\mathcal{A}$ is the set of actions, $\mathcal{T}: \mathcal{S} \times \mathcal{A} \times \mathcal{S} \rightarrow [0, 1]$ is the transition function, $\mathcal{R}: \mathcal{S} \rightarrow \mathbb{R}$ is the reward function, $\gamma \in [0, 1)$ is the discount factor, and $\mathcal{S}_0$ is the initial state distribution. We do not have access to the reward function $\mathcal{R}$ which we have to learn from human preferences. We use $s_t \in \mathcal{S}$ and $a_t \in \mathcal{A}$ to denote the state and action at time $t$. A trajectory, $\xi \in \Xi$, is a finite sequence of states and actions; $\xi = ((s_t, a_t)^T_{t=0})$ where $T$ is the time horizon of the environment. 

For autonomous routing, a graph structure is commonly used in which the nodes represent the locations and the edges represent the streets \citep{barnesmassively}. The set of states $\mathcal{S}$ is therefore the set of nodes, and the actions at each node are the outgoing edges. Our goal is to learn the human's reward function $R_H$ which is modeled as a linear combination of weights $w$ and features $\Phi$ of a trajectory, $R_H(\xi) = w^T\Phi(\xi)$. Learning $R_H$ thus simplifies to learning the weights $w$.

\textbf{Preference learning.} The objective of preference learning is to learn $w$ by querying a human for their preferences between pairs of trajectories. A preference query typically asks "Do you prefer trajectory $\xi_A$ or $\xi_B$?" \citep{biyik2018batch}. If a human prefers $\xi_A$ over $\xi_B$, it implies $R_H(\xi_A) > R_H(\xi_B)$, or equivalently $w^T\Phi(\xi_A) > w^T\Phi(\xi_B)$. From this strict inequality, we can derive that $w^T(\Phi(\xi_A)-\Phi(\xi_B)) > 0$. Let $\psi(\xi_A, \xi_B) = \Phi(\xi_A)-\Phi(\xi_B)$ denote the difference between the features of the two trajectories. The human's preference $I$ can then be encoded by $I = sign(w^T\psi)$.

The human input may be noisy due to uncertainty in their preferences, which can be modeled using Boltzmann rationality, where the likelihood of a preference (Eqn. \ref{eqn: likelihood}) is determined by a softmax function: 
\begin{align}
    P(I \mid \mathbf{w}) &= \begin{cases} 
    \frac{\exp(R_H(\xi_A))}{\exp(R_H(\xi_A)) + \exp(R_H(\xi_B))} & \text{if } I = +1 \\
    \frac{\exp(R_H(\xi_B))}{\exp(R_H(\xi_A)) + \exp(R_H(\xi_B))} & \text{if } I = -1
    \end{cases}
    \label{eqn: likelihood}
\end{align}

Let $p(w)$ be our current belief distribution of the reward weights. We can perform a Bayesian update to compute the posterior given human input $I$, $p(w | I) \propto p(I | w)p(w)$. For uniqueness, we constrain the norm of the reward weights such that $\lVert w \rVert_2 \leq 1$. Since $p(w)$ can have arbitrary shapes, we use an adaptive Metropolis algorithm \citep{bj1080222083} to learn the posterior distribution. Based on \citet{erdem2020asking}, the algorithm presents the human with a preference query and updates the belief distribution of $w$ until a fixed number of iterations is reached.

\textbf{Active Synthesis of Preference Queries.} To learn $w$ efficiently using minimal queries, active learning methods select preference queries $(\xi_A, \xi_B)$ that maximize information gain. This is equivalent to maximizing the mutual information between the query and the estimated weights $w$ \citep{erdem2020asking}. Our objective function $f$ is
\begin{equation}
    \max_{\xi_A, \xi_B} f(\xi_A, \xi_B) = \max_{\xi_A, \xi_B} H(\mathbf{w}) - \mathbb{E}_{I}[H(\mathbf{w} | I)]
\label{eqn: mutual information}
\end{equation}
where $H(w) = -\mathbb{E}_w[ \allowbreak log(p(w))]$ is the information entropy of the belief $p(w)$. This objective finds preference queries such that the difference between the entropy of the prior and the posterior is maximized. However, evaluating this objective is computationally expensive \citep{biyik2018batch} and often leads to suboptimal solutions. The next section discusses our approach of using counterfactual reasoning and environment design to more effectively generate trajectories that optimize this information gain objective.

\section{Technical Approach}

Active preference learning faces challenges in generating trajectories that optimize for information gain (Eqn. \ref{eqn: mutual information}), as the objective function involves a pair of trajectories as variables. This task is further complicated by the fact that the optimization is typically constrained to a single environment, which may not adequately represent the full feature space, resulting in learned rewards that often fail to generalize effectively. We discuss our approach of using counterfactual reasoning and environment design to address these issues.
  
\subsection{Counterfactual Reasoning} \label{sec: counterfactual}
Counterfactual reasoning explores different trajectories that could result if various reward weights were the true weights. We maintain a belief over the weights while estimating the human's reward function, where each sample of weights could lead to a different policy and consequently different trajectories when the policy is executed. This allows us to pose counterfactual questions, such as "what if reward $i$ is the true reward as opposed to reward $j$?" Let $w_i$ be an instance of reward weights sampled from our belief. We can train a policy $\pi_i$ that maximizes the reward function based on $w_i$ by using RL algorithms such as value iteration or PPO \citep{sutton2018reinforcement}. If we rollout the policy $\pi_i$ in an environment, we get a trajectory $\xi_i$ (\cref{alg: counterfactual} lines \ref{cr: train policy}-\ref{cr: rollout}). 

By sampling reward weights and generating trajectories, we construct a set of counterfactual trajectories that represent different human preferences. We then evaluate the information gain objective (\cref{eqn: mutual information}) for each pair of trajectories to identify the most informative preference query (\cref{alg: counterfactual} lines \ref{cr: infogain}-\ref{cr: max}). To minimize the number of evaluations of the objective function, we start by sampling $N$ reward weights. We then select the most diverse $M$ weights, where $M < N$, from this set by sequentially computing diversity based on cosine similarity, forming our final set of reward weights for evaluation (\cref{alg: counterfactual} lines \ref{cr: sample}-\ref{cr: diverse weights}).

\begin{algorithm}
\caption{Counterfactual Reasoning}
\begin{algorithmic}[1]
\Require Belief $P(w)$, $N$ samples, $M$ subset size

\State Sample $\{w_1, \dots, w_N\} \sim P(w)$ \label{cr: sample}
\State Select $M$ diverse weights (e.g., max cosine distance) \label{cr: diverse weights}
\For{each selected $w_k$}
    \State Train policy $\pi_k$ to maximize $w_k^T \Phi(\xi)$ \label{cr: train policy}
    \State Generate trajectory $\xi_k$ by rolling out policy $\pi_k$ \label{cr: rollout}
\EndFor
\State Compute information gain (Eq. \ref{eqn: mutual information}) for all pairs $\xi_i, \xi_j$  \label{cr: infogain}
\State Return most informative pair $(\xi_i, \xi_j)$ \label{cr: max}
\end{algorithmic}
\label{alg: counterfactual}
\end{algorithm}

\subsection{Environment Design} \label{sec: env design}

While counterfactual reasoning generates trajectory pairs optimizing for different reward weights, the fixed environment can limit their ability to reveal crucial preference distinctions. We posit that if we have the ability to "imagine" new environments or scenarios, we can better generate trajectories that show the differences between the different reward weights.

More formally, let $\Theta_E$ denote the set of environment parameters that the algorithm can modify (e.g. terrain type). The feature function $\Phi$ now depends on $\theta_E \in \Theta_E$ and can be represented as $\Phi_{\theta_E}$. The return of a trajectory is thus modified as $R_H(\xi) = w^T\Phi_{\theta_E}(\xi)$. Let $F$ be the information gain objective from Eqn. \ref{eqn: mutual information} but it includes $\theta_E$ as a parameter. We formulate environment design as a bilevel optimization problem:
\begin{equation}
    \max_{\theta_E, \xi_A, \xi_B} F(\xi_A, \xi_B, \theta_E) \quad \text{s.t.} \quad (\xi_A, \xi_B) \in \text{arg}\max_{\xi_A, \xi_B} f(\xi_A, \xi_B) \label{eqn: bilevel opt}
\end{equation}

The upper level optimization of objective function $F$ selects environment parameters $\theta_E$ which is then used by the lower level optimization to find the preference query $\xi_A$ and $\xi_B$ that maximizes information gain $f$ from Eqn. \ref{eqn: mutual information}.

Since $F(\xi_A, \xi_B, \theta_E)$ is generally not differentiable with respect to $\theta_E$, we use Bayesian optimization, a global optimization method that uses a Gaussian process (GP) to model $F$ \citep{snoek2012practical}. GP is a distribution on functions which has a mean function $m: \Theta_E \rightarrow \mathbb{R}$ and a positive definite covariance function $K: \Theta_E \times \Theta_E \rightarrow \mathbb{R}$. We use the upper confidence bound (UCB) as the acquisition function that selects the next $\theta_E$ to evaluate by finding $\theta_E$ that maximizes $UCB(\theta_E) = \mu(\theta_E) + \kappa \sigma(\theta_E)$. $\kappa$ is a hyperparameter that balances exploitation against exploration. Algorithm \ref{alg: environment design} shows the pseudocode for environment design.

\begin{algorithm}
\caption{Environment Design}
\begin{algorithmic}[1]
\Require Environment parameters $\Theta_E$, Bayesian optimization iterations $T$
\For{$t = 1$ to $T$} 
    \State Propose $\theta_E^t$ using Bayesian optimization
    \State Generate $(\xi_A, \xi_B)$ via CR (\Cref{alg: counterfactual}) in env $\theta_E^t$
    \State Compute information gain $F(\xi_A, \xi_B, \theta_E^t)$
    \State Update GP model with $(\theta_E^t, F(\xi_A, \xi_B, \theta_E^t))$
\EndFor
\State Return optimal $\theta_E^*$ found and corresponding $(\xi_A, \xi_B)$
\end{algorithmic}
\label{alg: environment design}
\end{algorithm}

\section{Experiments}

\begin{figure}[ht]
    \begin{subfigure}{0.65\linewidth}
      \centering
      \includegraphics[width=0.23\linewidth]{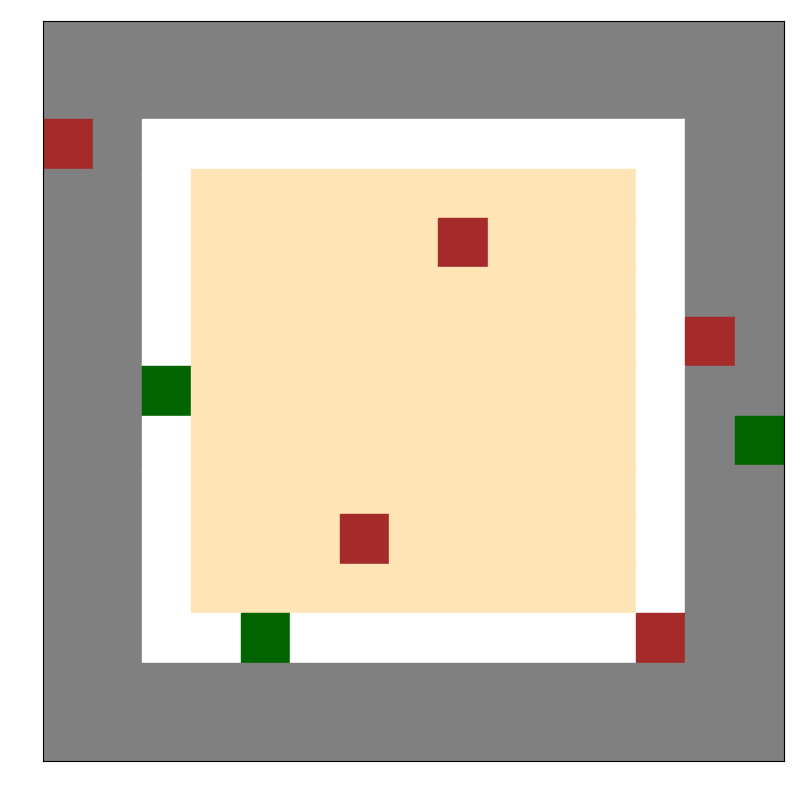}
      \includegraphics[width=0.23\linewidth]{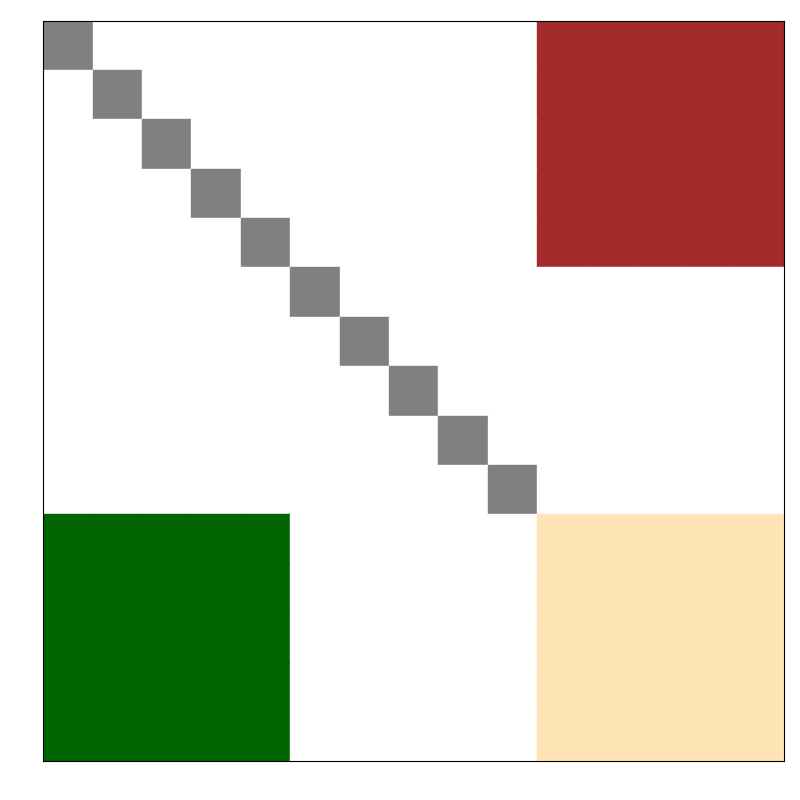}
      \includegraphics[width=0.23\linewidth]{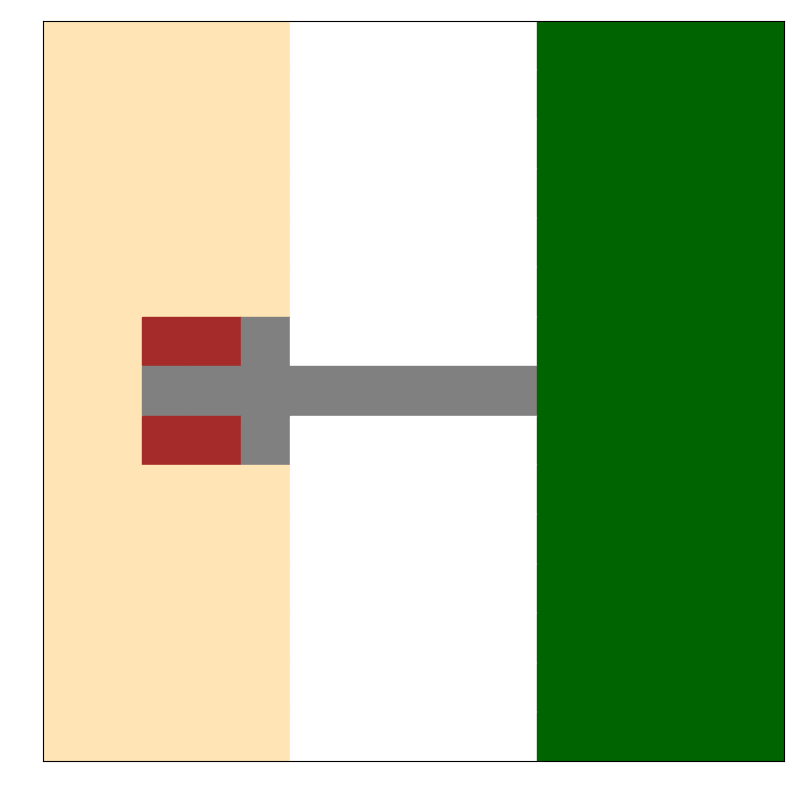}
      \includegraphics[width=0.23\linewidth]{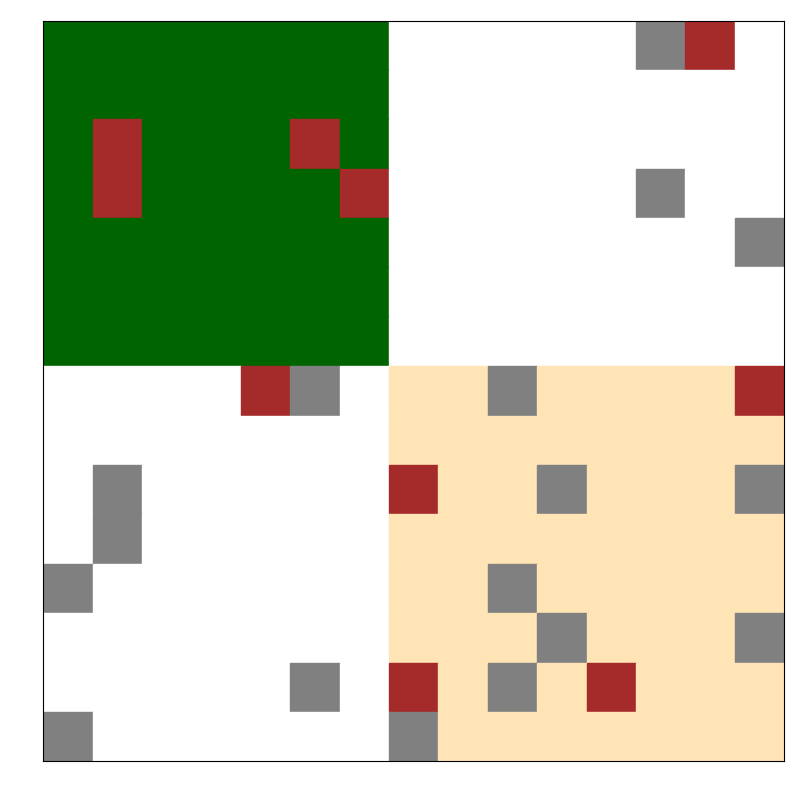}
      \caption{GridWorld Navigation}
      \label{fig:gridworld}
    \end{subfigure}%
    \begin{subfigure}{0.35\linewidth}
        \centering
        \includegraphics[width=0.4\linewidth]{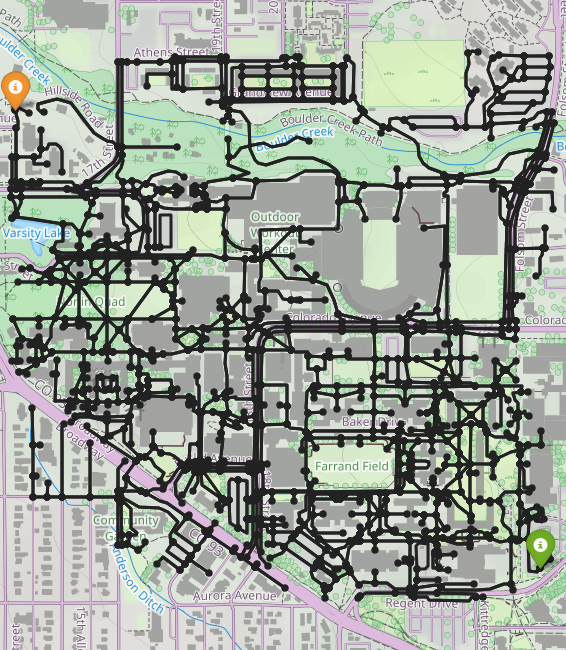}
        \includegraphics[width=0.4\linewidth]{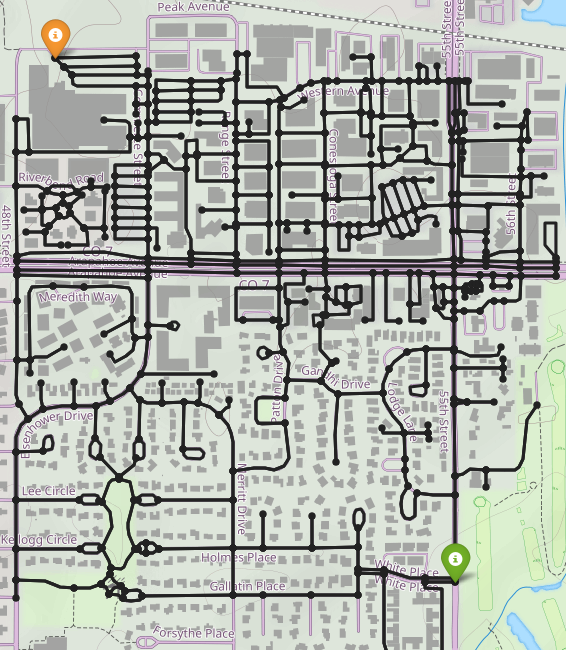}
        \caption{OpenStreetMaps}
        \label{fig:streetnav}
    \end{subfigure}
    \caption{(a) Sample environments used in the GridWorld Navigation experiments. The terrain types are brick (red), gravel (gray), sand (moccasin), and grass (green). (b) The street graph used in the OpenStreetMap Navigation experiments. Nodes and edges are highlighted in black.}
    \label{fig:experiment_domains}
\end{figure}

Our experiments aim to address the following hypotheses:
\begin{itemize}
    \item \textbf{H1:} CRED generates queries with higher information gain than those generated by baselines.
    \item \textbf{H2:} CRED learns more accurate rewards with fewer iterations compared to baselines.
    \item \textbf{H3:} CRED learns reward functions that generalize more effectively to novel environments.

\end{itemize}

\subsection{Environment Setup}

\textbf{GridWorld Navigation.} The GridWorld environment includes various terrains, such as brick, gravel, sand, and grass (Fig. \ref{fig:gridworld}). These environments were created by prompting GPT-4 to create realistic scenarios with a diverse distribution of features. From left to right, the environments were titled arid highlands, crossroads pass, coastal village, and forest desert. Our goal is to assess whether our approach effectively learns reward functions that capture the trade-offs between traversing different terrain types. We do not include time and safety as explicit features since they can be inferred from the terrains traversed. Each environment is a 15 x 15 grid where the goal is positioned in the bottom-right corner. We use arid highlands for training and the rest for testing. We also used different environments for training and observed similar results. Thus, we only present the results from training on arid highlands.

The number of environment parameters, $|\theta_E|$, corresponds to the total number of grid cells---225---which is too high for Bayesian optimization (Sec. \ref{sec: env design}) to perform efficiently. To address this, we use variational autoencoders or VAEs \citep{cinelli2021variational} to compress the distribution of environments into a lower-dimensional latent space $\mathcal{Z}$. By optimizing over this latent space, we can effectively learn the mutual information objective $F$ (Eqn. \ref{eqn: bilevel opt}) as a function of environment parameters.

\textbf{OpenStreetMap Navigation.} We also evaluate CRED on a real-world routing task using data from \citet{OpenStreetMap} (Fig. \ref{fig:streetnav}). The features considered include distance, travel time, and elevation changes (ascents/descents). Our experiments focus on last-mile deliveries, restricting the map area to a 500-meter radius from a central latitude and longitude point.

For training, we use a simplified street network consisting of 9 nodes and 12 edges, where distances, travel times, and elevations are sampled uniformly from the ranges [1, 5], [2, 5], and [-1, 1], respectively. In the test environments, these values range from [0.9, 405], [0.1, 291], and [-4.7, 4.7], respectively, though they are heavily skewed towards the lower end. This design ensures that the training differs from the testing environments, making it out of distribution. The environment parameters are the edge features, resulting in a total of 36 variables. While we initially experimented with variational graph autoencoders \citep{kipf2016variational} to optimize over a lower-dimensional latent space, we found that directly using edge features as environment parameters yielded better performance.

\subsection{Baselines}

We compare our approach to two state of the art preference learning algorithms. First, we optimize the mutual information objective (Eqn. \ref{eqn: mutual information}) over pre-generated trajectories obtained from random rollouts \citep{erdem2020asking} which we term \textbf{RR}. In addition, we adopt the trajectory generation method from \citet{christiano2017deep}. We train a policy using the mean of the current belief and generate trajectories through rollouts, allowing the policy to take random actions with probability $\epsilon$ ($25\%$ in our experiments). We refer to this baseline as the \textbf{M}ean \textbf{B}elief \textbf{P}olicy, abbreviated as \textbf{MBP}. Furthermore, we perform ablations of our full system: first, we combine our approach environment design (ED) (Sec. \ref{sec: env design}) with MBP and refer to this as \textbf{MBP + ED}. Lastly, we use counterfactual reasoning (Sec. \ref{sec: counterfactual}) alone as a baseline, referring to it as \textbf{CR}.

\subsection{Metrics}

\textbf{Belief Entropy.} This metric measures the uncertainty of the estimated rewards. The belief over reward weights is estimated by using Monte Carlo Markov Chain which generates likely samples from the distribution. To compute entropy, we first fit a probability function over these samples using Gaussian kernel density estimation (KDE), which approximates the probability density by using Gaussian kernels as weights \citep{scott2015multivariate}. We then create a grid covering the dimensions of the weight vector and compute the probability of each grid cell using KDE. The entropy is approximated as $H \approx \sum_i p(x_i)logp(x_i) \Delta V$ where $\Delta V$ is the volume of each grid cell and $p(x_i)$ is the KDE-evaluated density at grid point $x_i$.

\textbf{Difference in Rewards.} This metric quantifies the percentage difference in cumulative rewards compared to the ground truth. After training, we sample a set of reward weights from the learned belief and train a policy $\pi_{est}$ for each. We compute the percentage difference in rewards between the estimated policy evaluated under the true reward $(R_{est})$ and the ground truth policy $(R_{gt})$: $diff(w_{est}) = (R_{est} - R_{gt})/abs(R_{gt}) * 100$. We report the expected value by averaging over the belief distribution: $\sum_{w_{est}} p(w_{est}) diff(w_{est})$. 

\textbf{Policy Accuracy.} 
This metric quantifies the similarity in action selection between a policy trained on the estimated reward and the ground truth policy. For sampled weights from the belief, we train an estimated policy $\pi_{est}$. Accuracy is the proportion of states where the optimal action of $\pi_{est}$ matches that of the ground truth policy $\pi_{gt}$. We report the expected accuracy, weighted by the probability of each sampled weight.

\textbf{Jaccard Similarity.} This metric measures the overlap in visited states between trajectories generated by estimated policies and the ground truth policy. For each estimated policy $\pi_{est}$, we generate a trajectory $\xi_{est}$ by executing rollouts with deterministic actions. The Jaccard similarity $J(\xi_{est}, \xi_{gt})$ is the ratio of shared states to the total unique states: $J(\xi_{est}, \xi_{gt}) = |\xi_{est} \cap \xi_{gt}| / |\xi_{est} \cup \xi_{gt}|$. We report the expected Jaccard similarity, averaged over the belief distribution.

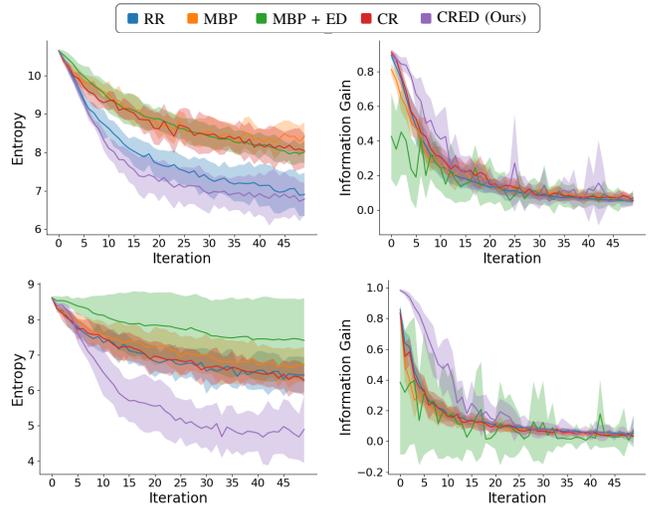
\begin{figure}
    \centering
    \begin{subfigure}{\linewidth}
        \centering
        \scalebox{0.6}{
        \begin{tikzpicture}       
            \node [label=right:{RR},fill={rgb,255:red,31; green,119; blue,180}, rounded corners=1pt] (node1) {};
            \node [label=right:{MBP},fill={rgb,255:red,255; green,127; blue,14}, rounded corners=1pt] (node2) at ([xshift=1.2cm]node1.east){};  
            \node [label=right:{MBP + ED},fill={rgb,255:red,44; green,160; blue,44}, rounded corners=1pt] (node3) at ([xshift=1.4cm]node2.east){};
            \node [label=right:{CR},fill={rgb,255:red,214; green,39; blue,40}, rounded corners=1pt] (node4) at ([xshift=2.2cm]node3.east){};
            \node [label=right:{CRED (Ours)},fill={rgb,255:red,148; green,103; blue,189}, rounded corners=1pt] (node4) at ([xshift=1.2cm]node4.east){};
            \draw[thick, rounded corners=3pt, gray] ($(node1.north west)+(-0.25,0.2)$) rectangle ($(node4.south east)+(2.4,-0.2)$);
        \end{tikzpicture}
        }
    \end{subfigure}
    \begin{subfigure}[b]{\linewidth} 
        \centering
        \begin{subfigure}[b]{0.48\textwidth}
            \includegraphics[width=\textwidth]{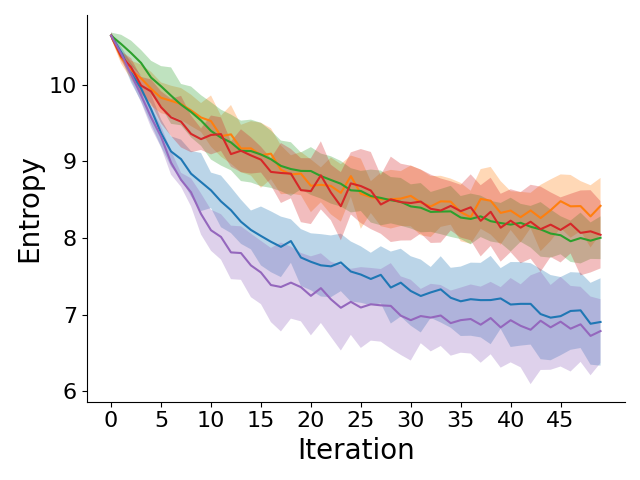}
        \end{subfigure}
        \begin{subfigure}[b]{0.48\textwidth}
            \includegraphics[width=\textwidth]{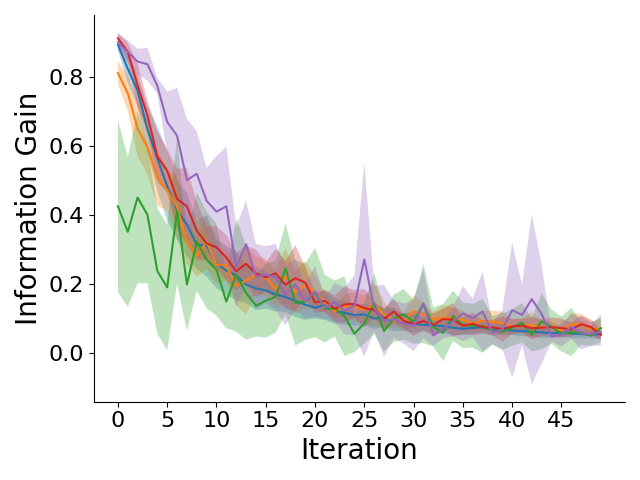}
        \end{subfigure}
        \begin{subfigure}[b]{0.48\textwidth}
            \includegraphics[width=\textwidth]{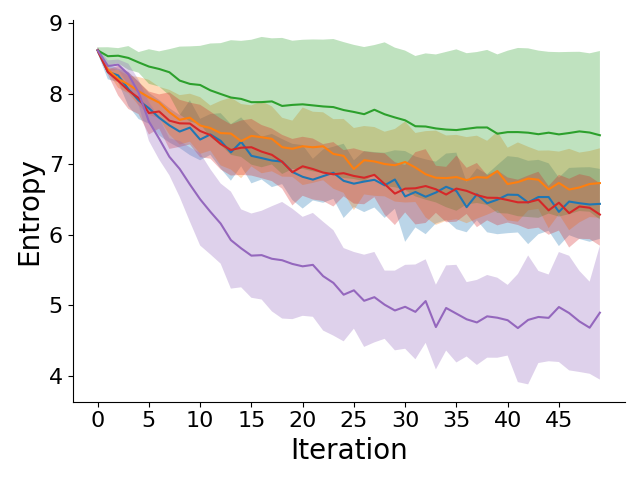}
        \end{subfigure}
        \begin{subfigure}[b]{0.48\textwidth}
            \includegraphics[width=\textwidth]{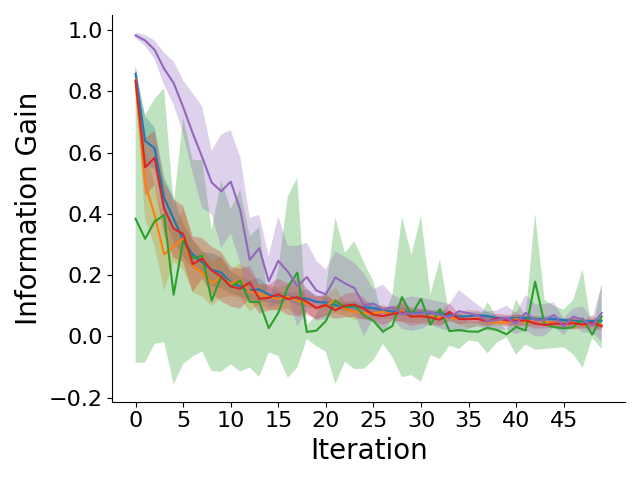}
        \end{subfigure}
    \end{subfigure}
    \caption{Belief entropy (left) and information gain from Eq.~\ref{eqn: mutual information} (right) are shown over the course of training for GridWorld (top) and OpenStreetMaps (bottom). Each line represents the mean, with the shaded region indicating the standard deviation. CRED selects preference queries with higher initial information gain, leading to lower entropy compared to the baselines.}
    \label{fig:entropy_plots}
\end{figure}

\section{Results}

\textbf{Information Gain of Preference Queries.} We evaluated our approach using 10 simulated users, each initialized with a different reward weight. This setup demonstrates our method's ability to learn a diverse range of reward functions. To ensure diversity among the ground truth weights, we sample 1000 random weight vectors and select the cluster centers obtained via K-Means as the ground truth weights. Figure \ref{fig:entropy_plots} shows the belief entropy and information gain objective across training iterations. In both experiments, CRED generates preference queries with higher information gain during the initial 10 iterations. As a result, the entropy of CRED decreases more rapidly in the early stages and ultimately converges to a lower value compared to the baselines, supporting \textbf{H1}.

\begin{figure*}[ht!]
    \centering

    \begin{subfigure}{\textwidth}
        \centering
        \renewcommand{\arraystretch}{1} 
        \begin{tabular}{@{}c@{} @{}c@{} @{}c@{}}  
            \hline
            {\footnotesize \quad \textbf{Difference in Rewards}} 
            & {\footnotesize \qquad \textbf{Policy Accuracy}}
            & {\footnotesize \textbf{Jaccard Similarity}}\\ 
            \hline
            \includegraphics[width=0.31\textwidth]{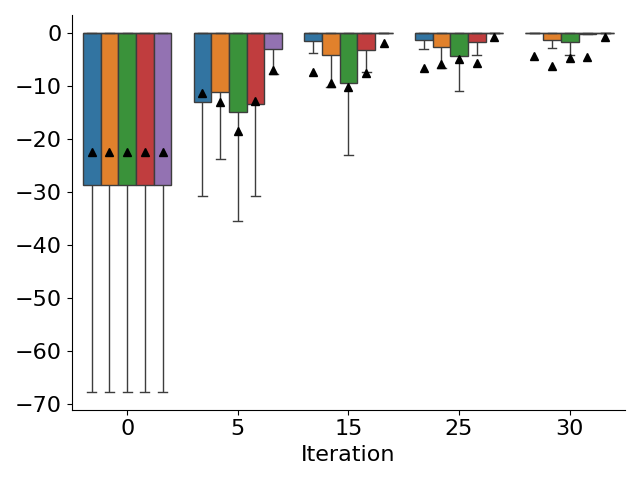} & 
            \includegraphics[width=0.31\textwidth]{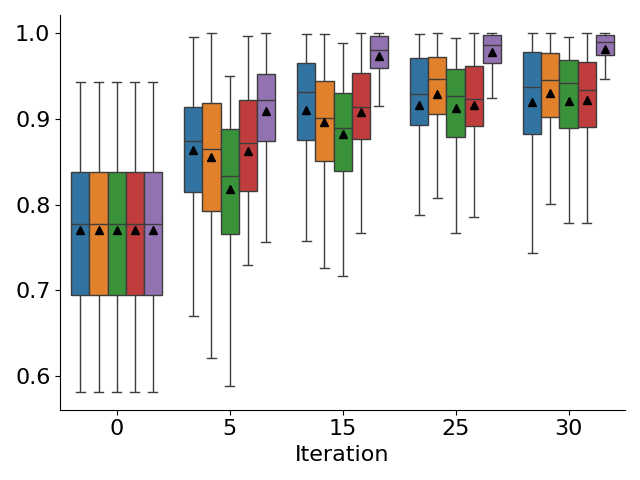} &
            \includegraphics[width=0.31\textwidth]{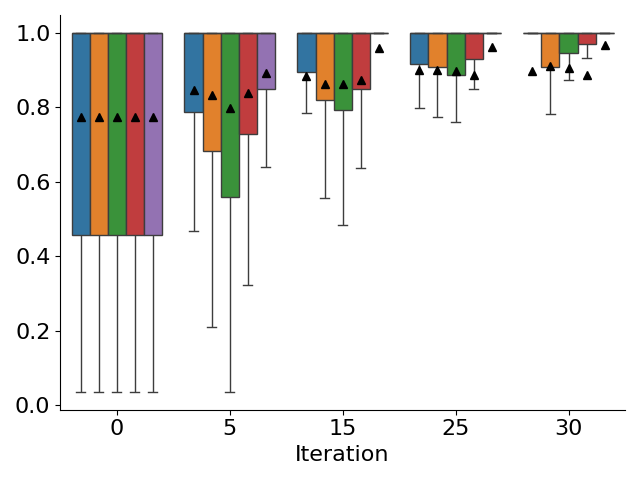} \\
        \end{tabular}
        \vspace{-7pt}
        \caption{GridWorld Navigation}
    \end{subfigure}

    \begin{subfigure}{\textwidth}
        \centering
        \renewcommand{\arraystretch}{1} 
        \begin{tabular}{@{}c@{} @{}c@{} @{}c@{}}  
            \hline
            \includegraphics[width=0.31\textwidth]{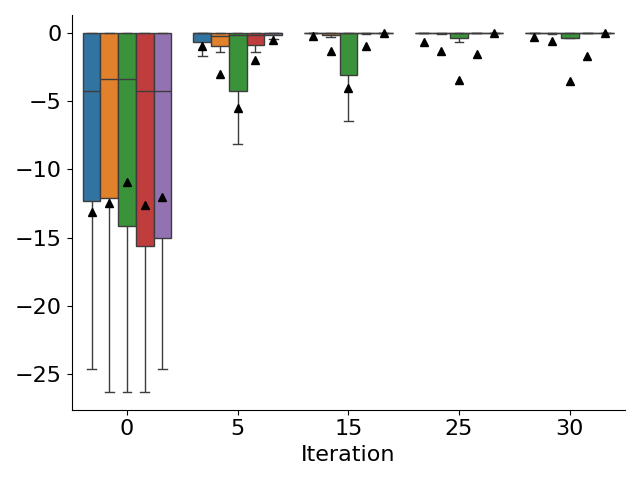} & 
            \includegraphics[width=0.31\textwidth]{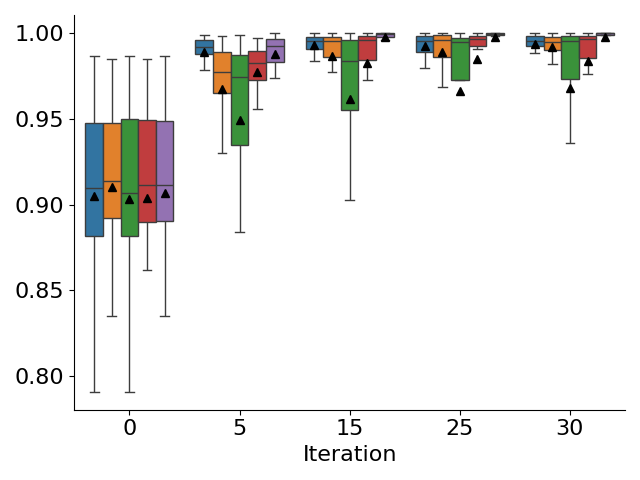} &
            \includegraphics[width=0.31\textwidth]{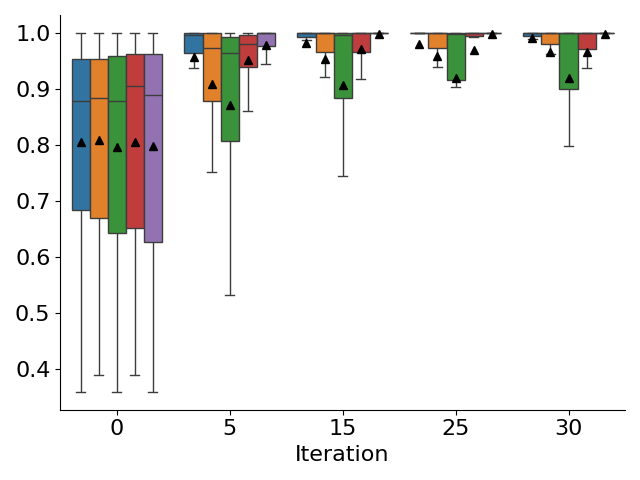} \\
        \end{tabular}
        \vspace{-7pt}
        \caption{OpenStreetMaps Navigation}
    \end{subfigure}
    \hfill
    \begin{subfigure}{\textwidth}
        \centering
        \scalebox{0.6}{
        \begin{tikzpicture}       
            \node [label=right:{RR},fill={rgb,255:red,31; green,119; blue,180}, rounded corners=1pt] (node1) {};
            \node [label=right:{MBP},fill={rgb,255:red,255; green,127; blue,14}, rounded corners=1pt] (node2) at ([xshift=1.2cm]node1.east){};  
            \node [label=right:{MBP + ED},fill={rgb,255:red,44; green,160; blue,44}, rounded corners=1pt] (node3) at ([xshift=1.4cm]node2.east){};
            \node [label=right:{CR},fill={rgb,255:red,214; green,39; blue,40}, rounded corners=1pt] (node4) at ([xshift=2.2cm]node3.east){};
            \node [label=right:{CRED (Ours)},fill={rgb,255:red,148; green,103; blue,189}, rounded corners=1pt] (node4) at ([xshift=1.2cm]node4.east){};
            \draw[thick, rounded corners=3pt, gray] ($(node1.north west)+(-0.25,0.2)$) rectangle ($(node4.south east)+(2.4,-0.2)$);
        \end{tikzpicture}
        }
    \end{subfigure}
    \caption{Box-and-whisker plots depicting the differences in rewards, policy accuracy, and Jaccard similarity when evaluating estimated rewards in the training and testing environments. Black triangles indicate the mean values. CRED converges to the ground truth more quickly and achieves higher performance in test environments.}
    \label{fig:policy results}
\end{figure*}
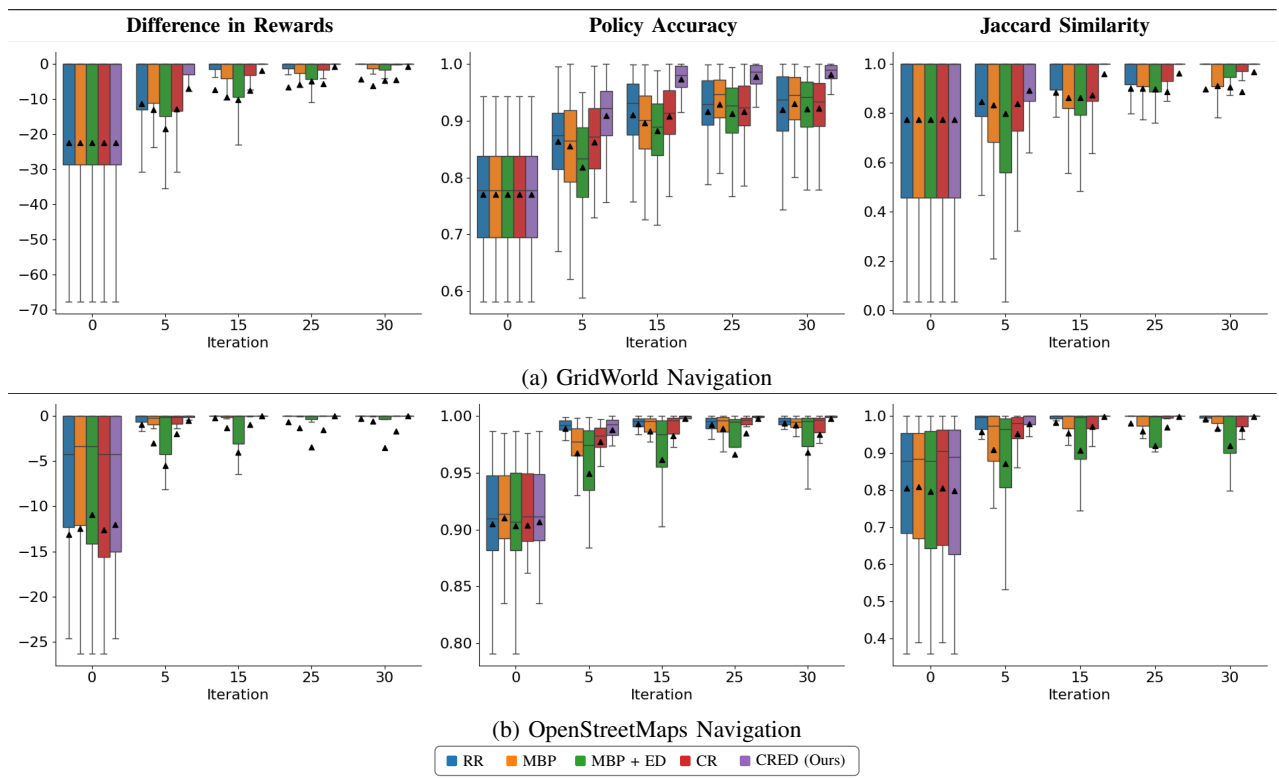




\begin{table*}[ht!]
    \centering
    \footnotesize
    \begin{tabularx}{\linewidth}{cXXX|XXX}
        \toprule
        & \multicolumn{3}{c}{\textbf{GridWorld}} & \multicolumn{3}{c}{\textbf{OpenStreetMaps}} \\
        \cmidrule(lr){2-4} \cmidrule(lr){5-7}
        \textbf{Condition} & \shortstack{\textbf{Diff. in} \\ \textbf{Rewards ($\uparrow$)}} & \shortstack{\textbf{Policy} \\ \textbf{Accuracy ($\uparrow$)}} & \shortstack{\textbf{Jaccard} \\ \textbf{Similarity ($\uparrow$)}}
        & \shortstack{\textbf{Diff. in} \\ \textbf{Rewards ($\uparrow$)}} & \shortstack{\textbf{Policy} \\ \textbf{Accuracy ($\uparrow$)}} & \shortstack{\textbf{Jaccard} \\ \textbf{Similarity ($\uparrow$)}} \\
        \midrule
        RR & -4.43 $\pm$ 13.49 & 0.92 $\pm$ 0.07 & 0.90 $\pm$ 0.22 & -0.33 $\pm$ 1.28 & 0.99 $\pm$ 0.01 & 0.99 $\pm$ 0.02 \\
        MBP & -6.32 $\pm$ 17.05 & 0.93 $\pm$ 0.06 & 0.91 $\pm$ 0.18 & -0.62 $\pm$ 2.35 & 0.99 $\pm$ 0.01 & 0.97 $\pm$ 0.11 \\
        MBP + ED & -4.77 $\pm$ 15.26 & 0.92 $\pm$ 0.06 & 0.91 $\pm$ 0.20 & -3.49 $\pm$ 10.62 & 0.97 $\pm$ 0.05 & 0.92 $\pm$ 0.14 \\
        CR & -4.54 $\pm$ 13.88 & 0.92 $\pm$ 0.07 & 0.89 $\pm$ 0.24 & -1.72 $\pm$ 7.49 & 0.98 $\pm$ 0.03 & 0.97 $\pm$ 0.07 \\
        CRED & \textbf{-0.70 $\pm$ 6.58*} & \textbf{0.98 $\pm$ 0.02*} & \textbf{0.97 $\pm$ 0.12*} & \textbf{-0.01 $\pm$ 0.05} & \textbf{1.00 $\pm$ 0.01} & \textbf{1.00 $\pm$ 0.00} \\
        \bottomrule
    \end{tabularx}
    \caption{Evaluation on test environments in the final training iteration. An asterisk (*) denotes statistical significance compared to all baselines based on a one-way ANOVA test.}
    \label{table:results}
\end{table*}

\textbf{Evaluation of Learned Rewards.} Figure \ref{fig:policy results} shows the box-and-whisker plots illustrating the difference in rewards, policy accuracy, and Jaccard similarity when evaluating the estimated rewards on the test environments. These plots show the distribution of the data, where the central line within each box represents the median, the upper and lower edges correspond to the first (Q1) and third (Q3) quartiles, and the whiskers extend to 1.5 times the interquartile range from Q1 and Q3. We also plot the mean as black triangles. 
In both experiments, CRED reaches the ground truth reward the fastest, converging within 15 iterations, supporting \textbf{H2}. The next best-performing algorithm, RR, takes approximately 25 iterations to converge when it does. Its relatively strong performance stems from the lack of a constraint requiring both trajectories in a preference query to reach the goal, allowing it to generate more diverse trajectories in the feature space. Table \ref{table:results} shows the mean and standard deviation of test environment metrics at the final training iteration. CRED achieves an $84\%$ and $97\%$ reduction in reward difference, a $5\%$ and $1\%$ absolute increase in policy accuracy, and a $6\%$ and $1\%$ absolute increase in Jaccard similarity compared to the best-performing baseline for GridWorld and OpenStreetMaps, respectively, supporting \textbf{H3}. Although the improvements in policy accuracy and Jaccard similarity for OpenStreetMaps are small, it's important to note that the initial values at iteration 0 were already high ($90\%$) due to a strong bias in the reward function towards reaching the goal. Even small differences in these percentages indicate substantial variations in how well preferences are followed.

\section{Conclusion}
\label{sec:conclusion}
In this work, we introduce CRED, a novel query generation method for active preference learning that uses counterfactual reasoning and environment design. By sampling from the current belief over reward weights, we pose the counterfactual "what if this reward were the true reward?" These counterfactuals generate trajectories that represent different human preferences. Meanwhile, environment design creates new scenarios to elicit human preferences in varied contexts, enabling the learned reward functions to generalize effectively to novel environments. Through experiments in GridWorld and OpenStreetMaps navigation, we demonstrate that CRED generates preference queries with higher information gain, learns human reward functions in fewer iterations, and achieves better generalization to novel environments compared to previous state of the art methods.

A limitation of our approach is that training a policy using reinforcement learning for the sampled reward weights can be time consuming. Potential solutions include parallelizing training or using meta-learning to develop a single policy that can be quickly fine-tuned for different reward weights.

\section*{Acknowledgments}
The authors would like to acknowledge Dusty Woods for assistance with figures and image editing.

\bibliographystyle{plainnat}
\bibliography{references} 

\end{document}